\documentclass[10pt,twocolumn,letterpaper]{article}

\usepackage{cvpr}      

\usepackage{xcolor}
\usepackage{algorithmic}
\usepackage{algorithm}
\usepackage[algo2e,linesnumbered,noend]{algorithm2e}
\usepackage{graphicx}
\usepackage{multirow}
\usepackage{amssymb}
\usepackage{epstopdf}
\usepackage{multicol}
\usepackage{tabularx}
\usepackage{multirow}
\usepackage{caption}
\usepackage{enumitem}
\usepackage{enumitem}
\usepackage{booktabs}
\usepackage{bm}
\usepackage{wrapfig}
\usepackage{amsmath}
\usepackage{pifont}
\usepackage{amssymb}
\usepackage{lipsum}
\usepackage{pifont}
\usepackage{amsfonts}
\usepackage{dsfont}
\usepackage{mathrsfs}
\usepackage{amsthm}
\usepackage{float}
\usepackage{multirow}
\usepackage{wrapfig}
\usepackage{makecell}

\usepackage[pagebackref,breaklinks,colorlinks]{hyperref}

\usepackage[capitalize]{cleveref}
\crefname{section}{Sec.}{Secs.}
\Crefname{section}{Section}{Sections}
\Crefname{table}{Table}{Tables}
\crefname{table}{Tab.}{Tabs.}

\def\cvprPaperID{8827} 
\def\confName{CVPR}
\def\confYear{2023}

\begin{document}

\title{BiCro: Modeling Relevance for Multi-modality Data by Bi-directional Cross-modal Similarity Matching}
\title{BiCro: Noisy Correspondence Rectification for Multi-modality Data via Bi-directional Cross-modal Similarity Consistency}


\author{Shuo~Yang\textsuperscript{\rm 1}\quad Zhaopan~Xu\textsuperscript{\rm 2}\quad Kai~Wang\textsuperscript{\rm 3}\quad Yang~You\textsuperscript{\rm 3}\quad\\ Hongxun~Yao\textsuperscript{\rm 2*}\quad Tongliang~Liu\textsuperscript{\rm 4}\quad Min~Xu\textsuperscript{\rm 1*}\\
		{\textsuperscript{\rm 1} University of Technology Sydney} 
		{\textsuperscript{\rm 2} Harbin Institute of Technology}\\
  {\textsuperscript{\rm 3} National University Singapore} 
  {\textsuperscript{\rm 4} The University of Sydney}\\
	}

\maketitle

\begin{abstract}
  As one of the most fundamental techniques in multimodal learning, cross-modal matching aims to project various sensory modalities into a shared feature space. To achieve this, massive and correctly aligned data pairs are required for model training. However, unlike unimodal datasets, multimodal datasets are extremely harder to collect and annotate precisely. As an alternative, the co-occurred data pairs (\textit{e.g.}, image-text pairs) collected from the Internet have been widely exploited in the area. Unfortunately, the cheaply collected dataset unavoidably contains many mismatched data pairs, which have been proven to be harmful to the model’s performance. To address this, we propose a general framework called BiCro (Bidirectional Cross-modal similarity consistency), which can be easily integrated into existing cross-modal matching models and improve their robustness against noisy data. Specifically, BiCro aims to estimate soft labels for noisy data pairs to reflect their true correspondence degree. The basic idea of BiCro is motivated by that -- taking image-text matching as an example -- \textit{similar} images should have \textit{similar} textual descriptions and vice versa. Then the consistency of these two \textit{similarities} can be recast as the estimated soft labels to train the matching model. The experiments on three popular cross-modal matching datasets demonstrate that our method significantly improves the noise-robustness of various matching models, and surpass the state-of-the-art by a clear margin.\looseness-1 The code is available at \url{https://github.com/xu5zhao/BiCro}.

\end{abstract}
\renewcommand{\thefootnote}{}
\footnotetext{*Corresponding authors: h.yao@hit.edu.cn; min.xu@uts.edu.au}
\section{Introduction}

As a key step towards general intelligence, multimodal learning aims to endow an agent with the ability to extract and understand information from various sensory modalities.
One of the most fundamental techniques in multimodal learning is cross-modal matching. Cross-modal matching tries to project different modalities into a shared feature space, and it has been successfully applied to many areas, \textit{e.g.,} image/video captioning~\cite{anderson2018bottom,li2019visual,li2017image}, cross-modal hashing~\cite{jiang2017deep,lee2018stacked}, and visual question answering~\cite{zhao2017video,kafle2017visual}.

\begin{figure}[!t]
\centering
  \resizebox{0.9\linewidth}{!} {
    \includegraphics{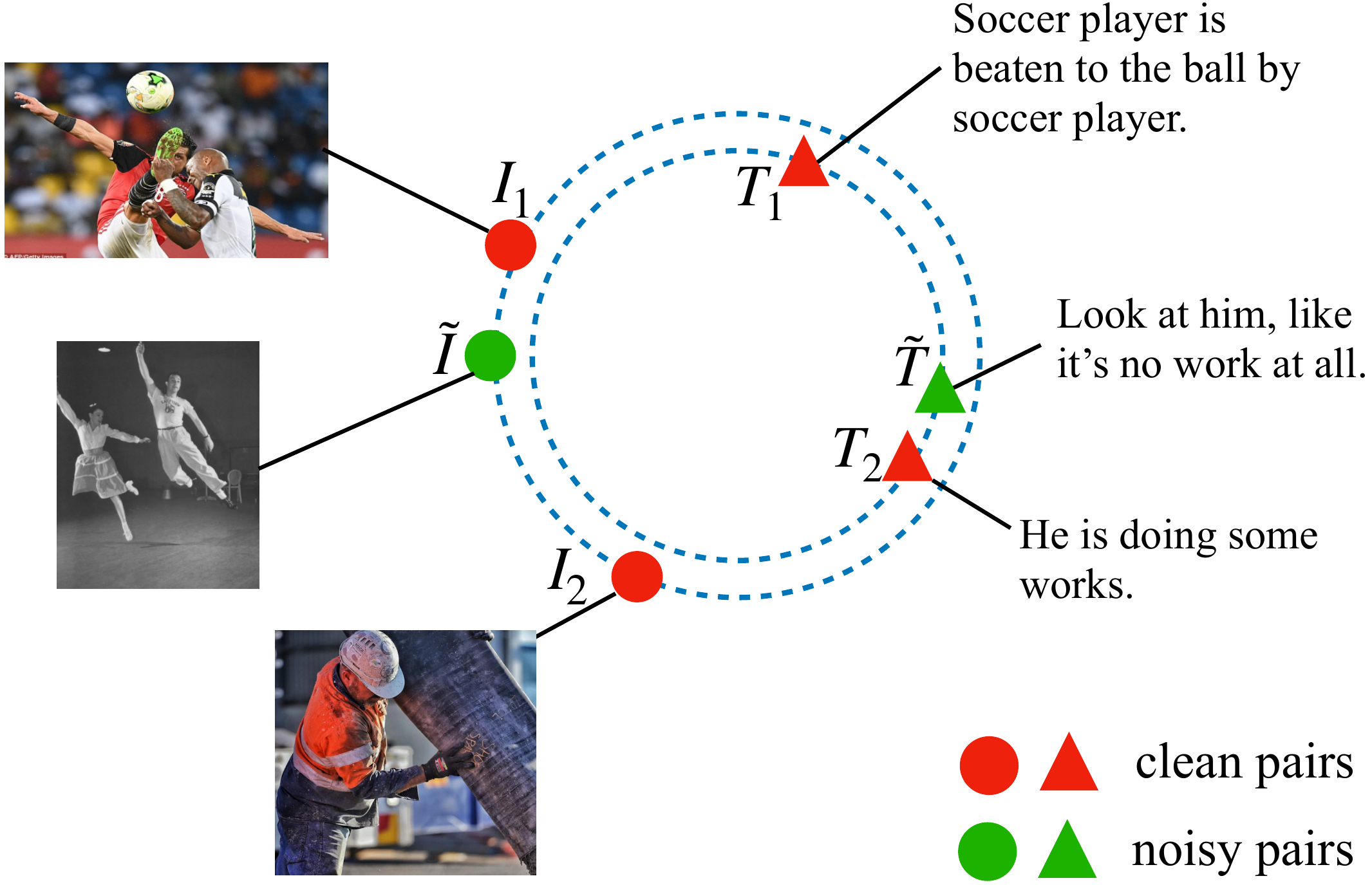}
  }
  \caption{Illustration of the \textit{Bidirectional Cross-modal similarity consistency (BiCro)}. In an image-text shared feature space, assume $(I_1,T_1)$ and $(I_2,T_2)$ are two clean positive data pairs. The image $\Tilde{I}$ is very close to the image $I_1$, but their corresponding texts $\Tilde{T}$ and $T_1$ are far away from each other. Similarly, the text $\Tilde{T}$ is very close to the text $T_2$ while their corresponding images $\Tilde{I}$ and $I_2$ are irrelevant. Therefore the data pair ($\tilde{I},\tilde{T}$) is possibly noisy (mismatched).}
    \label{fig:bicro}
  \vspace{-4mm}
\end{figure}

Like many other tasks in deep learning, cross-modal matching also requires massive and high-quality labeled training data. But even worse, the multimodal datasets (\textit{e.g.,} in image-text matching) are significantly harder to be manually annotated than those unimodal datasets (\textit{e.g.,} in image classification). This is because the textual description of an image is very subjective, while the categorical label of an image is easier to be determined. Alternatively, many works collect the co-occurred image-text pairs from the Internet to train the cross-modal matching model, \textit{e.g.}, CLIP~\cite{clip} was trained over 400 million image-text pairs collected from the Internet. Furthermore, one of the largest public datasets in cross-modal matching, Conceptual Caption~\cite{sharma2018conceptual}, was also harvested automatically from the Internet. Such cheap data collection methods inevitably introduce noise into the collected data, \textit{e.g.}, the Conceptual Caption dataset~\cite{sharma2018conceptual} contains $3\% \sim 20\%$ mismatched image-text pairs. 
The noisy data pairs undoubtedly hurt the generalization of the cross-modal matching models due to the memorization effect of deep neural networks~\cite{zhang2017understanding}.

\begin{figure*}[!t]
  \resizebox{\linewidth}{!} {
    \includegraphics{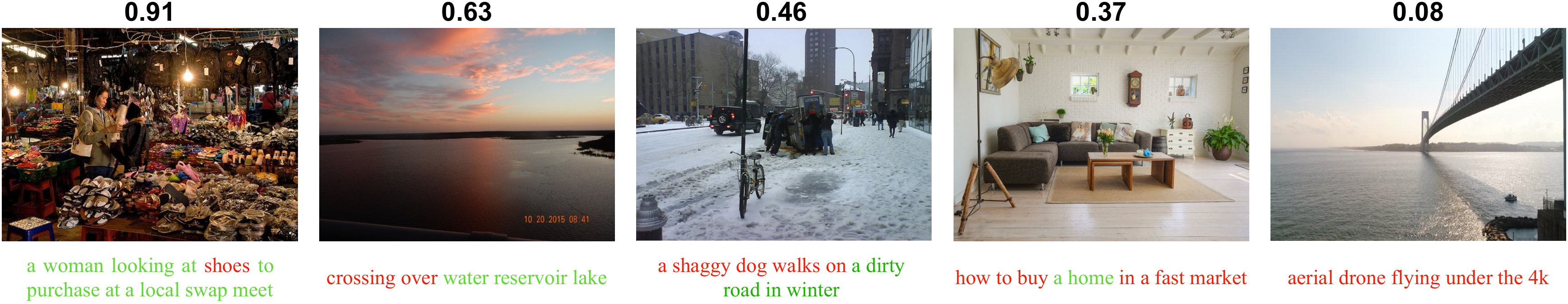}
  }
  \caption{Some noisy data pairs in the Flickr30K~\cite{young2014image} and Conceptual Caption~\cite{sharma2018conceptual} datasets. The first row shows our estimated soft correspondence labels for the image-text pairs illustrated in the second and third rows. We highlight the matched words in green and the mismatched words in red.}
    \label{fig:visualization}
  \vspace{-4mm}
\end{figure*}

Different from the noisy labels in classification, which refer to categorical annotation errors, the noisy labels in cross-modal matching refer to alignment errors in paired data. Therefore most existing noise-robust learning methods developed for classification cannot be directly applied to the cross-modal matching problem. Generally, all the data pairs in a noisily-collected cross-modal dataset can be categorized into the following three types based on their noise level: \textit{well-matched}, \textit{weakly-matched}, and \textit{mismatched}. The \textit{well-matched} data pairs can be treated as clean examples, and the \textit{weakly-matched} and \textit{mismatched} data pairs are noisy examples since they are all labeled as positive ($y=1$) in the dataset. We illustrate some noisy data pairs, including both \textit{weakly-matched} and \textit{mismatched} pairs in Fig.~\ref{fig:visualization}. The key challenge in noise-robust cross-modal matching is \textit{how to estimate accurate soft correspondence labels for those noisy data pairs.} The estimated soft label is expected to be able to depict the \textit{true correspondence degree} between different modalities in a data pair.

In this work, we propose a general framework for soft correspondence label estimation given only noisily-collected data pairs. Our framework is based on the \textit{Bidirectional Cross-modal similarity consistency (BiCro)}, which inherently exists in paired data. An intuitive explanation of the \textit{BiCro} is -- taking image-text matching task as an example -- similar images should have similar textual descriptions and vice versa. In other words, if two images are highly similar but their corresponding texts are irrelevant, there possibly exists image-text mismatch issue (see Fig.~\ref{fig:bicro}). Specifically, we first select a set of clean positive data pairs as \textit{anchor points} out of the given noisy data. This step is achieved by modeling the per-sample loss distribution of the dataset using a Beta-Mixture-Model (BMM) and selecting those samples with a high probability of being clean. Then, with the help of the selected anchor points, we compute the \textit{Bidirectional Cross-modal similarity consistency (BiCro)} scores for every noisy data pair as their soft correspondence labels. Finally, the estimated soft correspondence labels are recast as soft margins to train the cross-modal matching model in a co-teaching manner~\cite{han2018co} to avoid the self-sample-selection error accumulation problem. 

Before delving into details, we summarize our main contributions below:
\begin{itemize}
    \item This paper tackles a widely-exist but rarely-explored problem in cross-modal matching, \textit{i.e.,} the noisy correspondence issue. We identify and explore the key challenge of noise-robust cross-modal learning: how to rectify the \textit{binary noisy correspondence labels} into more accurate \textit{soft correspondence labels} for noisy data pairs.
    
    \item We propose a general framework called \textit{Bidirectional Cross-modal similarity consistency (BiCro)} for soft correspondence label estimation given only noisy data pairs. The BiCro is motivated by a rational assumption that -- taking the image-text matching as an example -- \textit{similar images should have similar textual descriptions and vice versa.} The BiCro can be easily adapted to any cross-modal matching models to improve their noise robustness. 
    
    \item To compute the BiCro score for noisy data pairs, we propose to first select some clean positive data pairs as \textit{anchor points} out of the noisy dataset by modeling the per-sample loss distribution of the dataset through a \textit{Beta-Mixture-Model (BMM)}. Then a set of examples with high clean possibilities can be selected. The BMM has a better modeling ability on the skewed clean data loss distribution compared to the Gaussian-Mixture-Model (GMM) used in the previous method~\cite{NCR}.
    
    \item Extensive experiments on three cross-modal matching datasets, including both synthetic and real-world noisy labels, demonstrate that our method surpasses the state-of-the-art by a large margin. The visualization results also indicate that our estimated soft correspondence labels are highly-aligned with the correlation degree between different modalities in data pairs.
\end{itemize}

\section{Related Works}
\subsection{Cross-modal Matching}
Cross-modal Matching is an essential research area of computer vision and multimedia, which aims at retrieving similar items that are of different modals with respect to the query modal. Image and text information are the most common modalities thus have been widely developed. The main challenge of this task is to measure the similarity between different modalities. Most existing works map images and texts into a shared latent space where they can be aligned. According to the alignment, existing works could be divided into two categories: 1) Global alignment, the image and text are matched from a global feature computed by multiple neural networks. For example, \cite{WangLHL19,faghri2017vse++,Sarafianos0K19,ZhangLZL20} propose a two-stream global feature learning network and compute the pairwise similarity between global features. 2) Local alignment, the regions within an image and words in a sentence correspond with each other such that finer-grained detail is captured by mapping semantically similar items. For example, \cite{ChenLYK0G0020,KimSK21} adopt attention mechanism to explore the semantic region-word correspondences. \cite{li2019visual,DiaoZML21} infer relation-aware similarities with both the local and global alignments by the graph convolutional network. 
However, all these alignment methods assume that the training data pairs are perfectly aligned, which is impossible to satisfy due to the high collection and annotation cost.

\subsection{Learning with Noisy Labels}
Supervised training of deep learning models requires precisely labeled datasets and noisy labels can significantly degrade the generalization of models. Therefore, various methods were proposed to improve robustness against noisy labels.
The typical algorithms for combating noisy labels including adding regularization~\cite{han2018masking,guo2018curriculumnet,veit2017learning,vahdat2017toward,li2017learning,li2019gradient,wu2020class2simi}, designing robust loss functions~\cite{liu2016classification,xu2019l_dmi,patrini2017making,APL,xia2021robust,xia2019anchor,xia2019revision}, selecting possible clean samples~\cite{yao2020searching,yu2019does,han2018co,malach2017decoupling,ren2018learning,jiang2018mentornet,xia2021sample} and correcting the labels~\cite{ma2018dimensionality,tanaka2018joint,wang2022reliable,yang2022estimating,yang2021estimating}. We mainly introduce the label correction approaches which are most related to this work. Specifically, label correction algorithms aim to identify and correct suspicious labels in an iterative framework. Bootstrapping \cite{ReedLASER14} combines the noisy label and the DNN prediction as a soft pseudo-label to replace the original label. Furthermore, Dynamic bootstrapping \cite{ArazoOAOM19} use a Beta-Mixture-Model to combine label dynamically for every sample. Differently, Joint Optimization \cite{TanakaIYA18} trained their model with a large learning rate for several epochs firstly, and then generated pseudo-labels by averaging the predictions of that epochs. Recently, SELFIE~\cite{SongK019} and AdaCorr~\cite{ZhengWG0MC20} selectively combine the labels of noisy data for corresponding to true labels with a high probability. However, almost all existing noisy label studies, focusing on the classification task, cannot be adapted to cross-modal matching directly in which the noisy label refers to the alignment errors between data pairs rather than the categorical labeling errors. The label of cross-modal matching can be consider as $y_i \in \{0,1\}$, where data pair is positively correlated ($y_i = 1$) or not ($y_i = 0$). Therefore, label correction of noisy multi-modal data pairs can be translated into generating a continuous value between 0 and 1, called soft correspondence labels in this paper, to measure the true degree of correspondence. In this view, the methods proposed by NCR~\cite{NCR} and DECL~\cite{QinP00022} both improve their noise-robustness by correcting the soft correspondence label. Specifically, NCR~\cite{NCR} recasts the network's prediction as the estimated soft correspondence label and DECL~\cite{QinP00022} models the uncertainty of cross-modal correspondence to predict correct correspondence of paired data. However, these methods rely on the network's predictions, which would cause severe confirmation bias: Those confident but wrong predictions would be used to guide subsequent training, leading to a loop of self-reinforcing errors~\cite{chen2021two}. Differently, we rectify the noisy correspondence labels by the Bidirectional Cross-modal similarity consistency (BiCro) that is inherently contained in the paired data. Our soft correspondence labels are generated by considering the characteristics of multi-modal data itself, so as to avoid the confirmation bias problem of previous methods.

\section{Methodology}
\subsection{Problem Definition}\label{sec:def}
We first introduce the cross-modal matching task by taking image-text matching as an example. Then, we will present the noisy correspondence problem in cross-modal matching. 

Traditionally, given a dataset $\mathcal{D} = \{ (I_i, T_i, y_i)\}_{i=1}^N$, where $N$ indicates the number of training samples, $(I_i, T_i)$ is an image-text pair and $y_i \in \{0,1\}$ is the label. The binary label $y_i$ is a hard-labeled correspondence score which indicates that the pair $(I_i, T_i)$ is positively correlated ($y_i = 1$) or not ($y_i = 0$). The aim of cross-modal matching is to project the two modalities (visual and textual modalities in our case) into a shared feature space wherein positive data pairs have higher feature similarities and negative data pairs have lower feature similarities. Generally, the similarity of the given image-text pairs can be computed by $S(f(I),g(T))$, where $S$ is a similarity measurement, $f$ and $g$ are two modal-specific feature extractors. In the following, we will denote the $S(f(I),g(T))$ as $S(I,T)$ for symbol simplicity. The feature extractors $f$ and $g$ can be learned by minimizing the following triplet loss:
\begin{align}\label{eq:triplet_loss}
L_{hard}\left(I_i, T_i\right)&=\left[\alpha-S\left(I_i, T_i\right)+S\left(I_i, \hat{T}_h\right)\right]_{+}\nonumber\\
&+\left[\alpha-S\left(I_i, T_i\right)+S\left(\hat{I}_h, T_i\right)\right]_{+}
\end{align}

where $(I_i,T_i)$ is a positive data pair, $\alpha>0$ denotes a given margin, $[x]_{+}=\max (x, 0)$. $\hat{I}_h$ and $\hat{T}_h$ are hard negative samples~\cite{faghri2017vse++} (the most similar negative samples in a mini-batch, \textit{i.e.}, $\hat{I}_h=\operatorname{argmax}_{I_j \neq I_i} S\left(I_j, T_i\right)$ and $\hat{T}_h=\operatorname{argmax}_{T_i \neq T_i} S\left(I_i, T_j\right)$). 

The solution in Eq.~\ref{eq:triplet_loss} is based on the assumption that all collected data pairs $(I,T)$ are perfectly aligned. However, in practice, the multi-modal dataset is usually annotated by using cheap methods or harvested from the internet. Therefore, an unknown portion of data in the noisily-collected dataset $\Tilde{\mathcal{D}}=\{ (I_i, T_i, \Tilde{y}_i)\}_{i=1}^N$ is \textit{mis-labeled}, which means that some data pairs $(I, T)$ are \textit{mis-matched} or \textit{weakly-matched} but they are wrongly labeled as $\Tilde{y}=1$. Minimizing Eq.~\ref{eq:triplet_loss} on $\Tilde{\mathcal{D}}$ would result in a poorly-generalized cross-modal matching model since it would overfit to the noisy dataset and pulls close those negative samples.


\subsection{Robust Matching Loss by Softening the Correspondence Label}\label{sec:soft_loss}
As we have discussed in Section.~\ref{sec:def}, in a noisily-labeled dataset $\tilde{\mathcal{D}}$, the hard correspondence labels $\Tilde{y}$ are unreliable and cannot accurately reflect the degree of correlation between different modalities. Therefore, the noisy labels $\Tilde{y}$ need to be rectified to a more accurate estimation of the soft correspondence scores ${y^*} \in [0,1]$ between the modalities $I$ and $T$. The rectified soft labels ${y^*}$ are expected to be able to well-depict the correspondence degree between $I$ and $T$ (\textit{i.e.,} ${y^*}$ gradually grow from 0 to 1 as the correlation of $I$ and $T$ increases). Then, the rectified soft labels ${y^*}$ can be recast as the soft margin to learn the shared feature space by:

\begin{align}\label{eq:soft_loss}
L_{soft}\left(I_i, T_i\right)&=\left[\hat{\alpha}_i-S\left(I_i, T_i\right)+S\left(I_i, \hat{T}_h\right)\right]_{+}\nonumber\\
&+\left[\hat{\alpha}_i-S\left(I_i, T_i\right)+S\left(\hat{I}_h, T_i\right)\right]_{+}
\end{align}
where the $\hat{\alpha}_i$ is the soft margin determined by $y_i^*$, \textit{i.e.,} $\hat{\alpha}_i=\frac{m^{y^*_i}-1}{m-1} \alpha$ and $m$ is a hyper-parameter~\cite{NCR}, $\hat{I}_h$ and $\hat{T}_h$ are hard negative samples.

Now, the most challenging problem is: \textit{how to estimate accurate soft correspondence labels $y^*$ by only using the hard-labeled noisy dataset $\Tilde{\mathcal{D}}=\{ (I_i, T_i, \Tilde{y}_i)\}_{i=1}^N$}. NCR~\cite{NCR} proposes to leverage the network predictions to assign pseudo labels for every data pair. However, this would cause severe confirmation bias: Those confident but wrong predictions would be used to guide subsequent training, leading to a loop of self-reinforcing errors~\cite{chen2021two}. In this work, we propose to estimate the correspondence labels by leveraging the inherent similarity relationships between the two modalities, which gets rid of the confirmation bias problem.

\subsection{Soft Correspondence Label Estimation}

The key idea of our method is based on the assumption that \textit{similar images should have similar textual descriptions and vice versa}. Assuming we have collected some clean positive data pairs (their soft correspondence labels $y^*$ are 1), we can use these collected \textit{anchor points} to infer the soft correspondence labels for the rest noisy data pairs, based on the aforementioned assumption. We will first show how to collect the anchor points out of a given noisy dataset, then we will present how to infer the soft correspondence labels for the noisy data pairs by using the anchor points.
\begin{figure}[!t]
\centering
  \resizebox{0.9\linewidth}{!} {
    \includegraphics{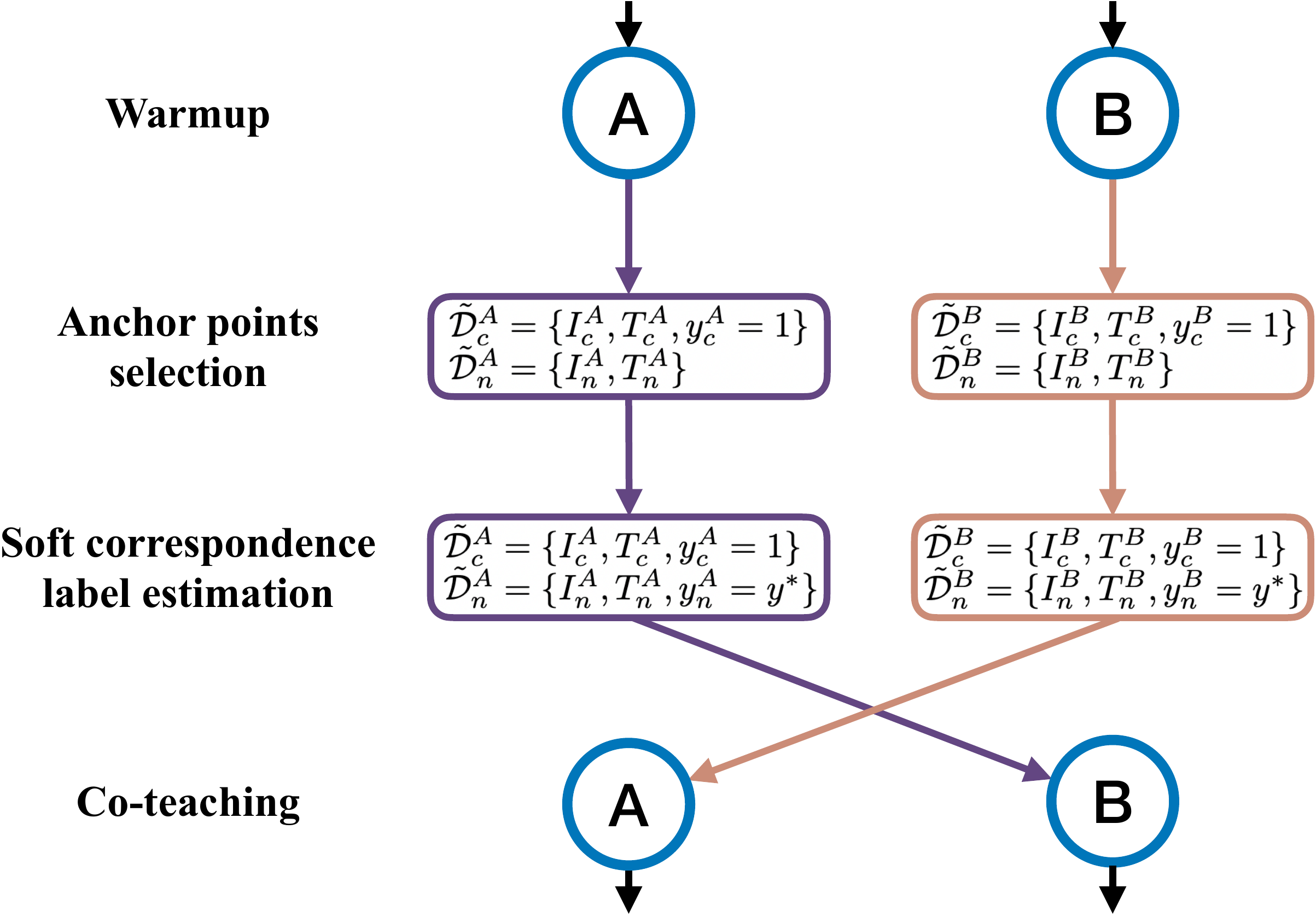}
  }
  \caption{The overall training pipeline. We simultaneously train two models $A$ and $B$ in a co-teaching~\cite{han2018co} manner to avoid the self-sample-selection error accumulation. At the inference stage, we average the predictions of $A$ and $B$.}
    \label{fig:training}
    \vspace{-4mm}
\end{figure}
\subsubsection{Anchor Points Selection by Modeling Per-sample Loss Value Distribution}\label{sec:anchor_point}
\begin{figure}[!t]
  \resizebox{\linewidth}{!} {
    \includegraphics{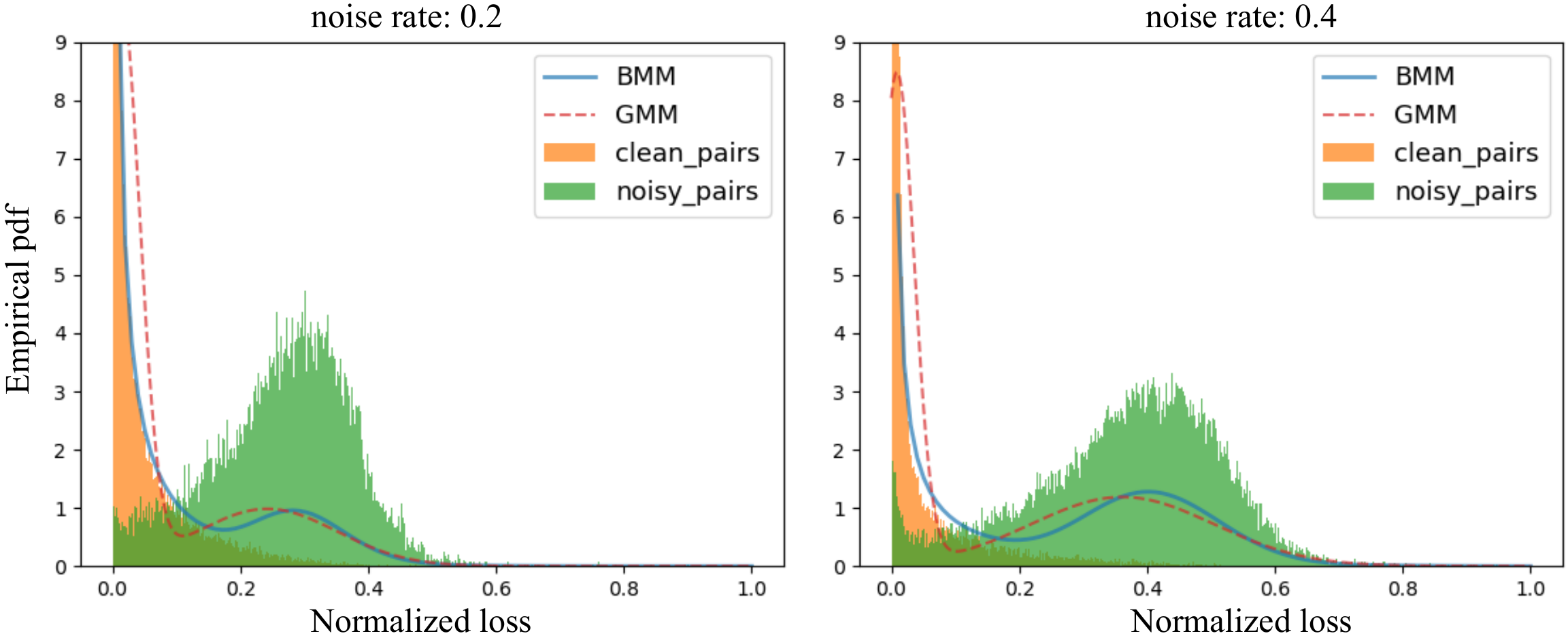}
  }
  \caption{The per-sample loss distribution under 20\% noisy labels and 40\% noisy labels. Beta-Mixture-Model (BMM) performs better in modeling the skewed clean set loss distribution than Gaussian-Mixture-Model (GMM).}
  \vspace{-5mm}
    \label{fig:bmm}
\end{figure}

We aim to identify those clean samples as anchor points in a noisy dataset $\Tilde{\mathcal{D}}$. The memorization effect of deep neural networks~\cite{zhang2021understanding} reveals that DNNs would first memorize training data of clean labels then those of noisy labels. This phenomenon indicates that noisy examples would have higher loss values while clean ones would have lower loss values during the early epochs of training. Therefore, given a matching model $(f,g,S)$, we can first compute the per-sample loss by:
\begin{equation}
\ell_{(f, g, S)}=\left\{\ell_i\right\}_{i=1}^N=\left\{L_{hard}\left(I_i, T_i\right)\right\}_{i=1}^N
\end{equation}
Then, we can utilize the difference of the per-sample loss value distribution to identify clean data pairs. NCR~\cite{NCR} adopts a two-component Gaussian Mixture Model to fit the per-sample loss value distribution:
\begin{equation}\label{eq:gmm}
p(\ell)=\sum_{k=1}^K \lambda_k p(\ell \mid k)
\end{equation}
where $\lambda_k$ and $p(\ell | k)$ are the mixture coefficient and the probability density of the $k$-th component, respectively. However, we empirically found the clean set loss distribution is extremely skewed towards zero (see Fig.~\ref{fig:bmm}), making the Gaussian a poor approximation to the clean set loss distribution. As a more flexible distribution, the beta distribution has a better modeling ability for the symmetric and skewed distributions. We propose to fit a beta mixture model (BMM) to approximate the loss distribution for mixtures of clean and noisy samples (see Fig.~\ref{fig:bmm}). The beta distribution over a normalized loss $\ell \in [0,1]$ is defined to have pdf:
\begin{equation}\label{eq:beta_distribution}
p(\ell \mid \gamma, \beta)=\frac{\Gamma(\gamma+\beta)}{\Gamma(\gamma) \Gamma(\beta)} \ell^{\gamma-1}(1-\ell)^{\beta-1}
\end{equation}
where $\gamma,\beta > 0$ and $\Gamma(\cdot)$ is the Gamma function. The mixture pdf can be formulated by substituting the Eq.~\ref{eq:beta_distribution} into Eq.~\ref{eq:gmm}. We use an Expectation Maximization (EM) procedure to fit the BMM to the observed per-sample loss value distribution. Then, we can compute the probability of the $i$-th data pair being clean or noisy as:
\begin{equation}\label{eq:clean_posterior}
p\left(k \mid \ell_i\right)=\frac{p(k) p\left(\ell_i \mid k\right)}{p\left(\ell_i\right)}
\end{equation}
where $k=0/1$ denotes the data pair $(I_i,T_i,y_i)$ is clean/noisy. Now we can select a set of anchor points $\Tilde{\mathcal{D}}_c = \{I_c,T_c,y_c = 1 \}$ from the noisy dataset $\Tilde{\mathcal{D}}$:
\begin{equation}\label{eq:anchor_select}
    \Tilde{\mathcal{D}}_c = \{(I_i,T_i,y_i=1) \mid p\left(k=0 \mid \ell_i\right) > \delta, \forall (I_i,T_i) \in \Tilde{\mathcal{D}}\}
\end{equation}
where $\delta$ is a threshold. Since the remained data $\Tilde{\mathcal{D}}_n = \Tilde{\mathcal{D}} \setminus \Tilde{\mathcal{D}}_c$ is possibly noisy, we drop their labels as:
\begin{equation}\label{eq:noisy_select}
    \Tilde{\mathcal{D}}_n = \{(I_i,T_i) \mid p\left(k=0 \mid \ell_i\right) \leq \delta, \forall (I_i,T_i) \in \Tilde{\mathcal{D}}\}
\end{equation}
In Section.~\ref{sec:label_est}, we will show how to estimate soft correspondence labels for $ \Tilde{\mathcal{D}}_n$ by using the anchor point samples in $ \Tilde{\mathcal{D}}_c$. Then, the collected clean data and the noisy data with their estimated soft labels can be used for the model training.

\begin{algorithm*}[th!]
\KwIn{A noisily-labeled dataset $\tilde{\mathcal{D}}$}
{
\textbf{Required}: the clean probability threshold $\delta$, two individual matching models $A$ and $B$ with different initializations and data batch sequences}

Warm up the model $(A,B)$ using $L_{hard}$ Eq.~\ref{eq:triplet_loss}.

\For{$i=1:num\_epochs$}
{
\textcolor{blue}{//Section.~\ref{sec:anchor_point}: modeling per-sample loss distribution using Eq.~\ref{eq:clean_posterior}}

$\mathcal{P}^A = \{p_i^A \mid p_i^A = p\left(k=0 \mid \ell_i\right)\}_{i=1}^N \leftarrow BetaMixtureModel(\tilde{\mathcal{D}},B)$

$\mathcal{P}^B = \{p_i^B \mid p_i^B = p\left(k=0 \mid \ell_i\right)\}_{i=1}^N \leftarrow BetaMixtureModel(\tilde{\mathcal{D}},A)$

\For{$k \in \{A,B\}$}
{
$\Tilde{\mathcal{D}}^k_c = \{(I_i,T_i,y_i=1) \mid p_i^k > \delta, \forall (I_i,T_i) \in \Tilde{\mathcal{D}}\}$ \hfill \textcolor{blue}{//Section.~\ref{sec:anchor_point}: anchor points selection using Eq.~\ref{eq:anchor_select}}

$\Tilde{\mathcal{D}}^k_n = \{(I_i,T_i) \mid p_i^k \leq \delta, \forall (I_i,T_i) \in \Tilde{\mathcal{D}}\}$\hfill \textcolor{blue}{//Section.~\ref{sec:anchor_point}: noisy data selection using Eq.~\ref{eq:noisy_select}}

\For{$j=1:num\_steps$}
{
Sample a mini-batch $\{\mathcal{B}_j^c = (I_c,T_c,y_c=1), \mathcal{B}_j^n= (I_n,T_n)\}$ from $\{\Tilde{\mathcal{D}}^k_c,\Tilde{\mathcal{D}}^k_n\}$

\textcolor{blue}{//Section.~\ref{sec:label_est}: soft label estimation by bidirectional cross-modal similarity consistency}

Keep the labels in $\mathcal{B}_j^c$, estimate soft correspondence labels $y^*$ for the noisy data $\mathcal{B}_j^n$ using Eq.~\ref{eq:soft_label}

\textcolor{blue}{//Section.~\ref{sec:soft_loss}: optimize the soft matching loss on the estimated soft correspondence label }

Train the network $k$ on $\{\mathcal{B}_j^c = (I_c,T_c,y_c=1), \mathcal{B}_j^n= (I_n,T_n,y_n=y^*)\}$ by minimizing Eq.~\ref{eq:soft_loss}
}
}
}

\KwOut{Matching models $(A,B)$}
\caption{The training pipeline of our robust cross-model matching framework.}
\label{algo}
\end{algorithm*}

However, training a model with high-confident (low loss) examples selected by the model itself would cause severe error accumulation problem, which is widely acknowledged in noisy label learning~\cite{han2018co,yu2019does,li2020dividemix}. Similar to NCR~\cite{NCR}, we also adopt the co-teaching~\cite{han2018co} paradigm to alleviate the error accumulation problem. Specifically, we simultaneously train two networks $A = \{ f^A,g^A,S^A\}$ and  $B = \{ f^B,g^B,S^B\}$ with the same architecture but different data sequences and initializations. At each training epoch, the network $A$ will model the per-sample loss distribution on its own data batch sequence $\Tilde{\mathcal{D}}^A$ and (1) select the anchor points $\Tilde{\mathcal{D}}^A_c = \{I^A_c,T^A_c,y^A_c = 1 \}$ according to Eq.~\ref{eq:anchor_select}, (2) estimate the soft correspondence labels for the noisy data $\Tilde{\mathcal{D}}^A_n = \{I^A_n,T^A_n, y^A_n = y^*\}$, then the data $\Tilde{\mathcal{D}}^A_c \cup \Tilde{\mathcal{D}}^A_n$ is used for training the other network (\textit{i.e.,} B). Note that, before the co-teaching procedure, a warmup process is conducted to make the network $A$ and $B$ achieve an initial convergence. In order to reduce the impact of mismatched data pairs in the warmup process, we only select a certain ratio (\textit{i.e.,} warmup selection ratio $\alpha$) of the data pairs with smaller loss values for training in each batch. The training pipeline is illustrated in Fig.~\ref{fig:training}.


\subsubsection{Soft Correspondence Label Estimation by Cross-modal Similarity Consistency}\label{sec:label_est}
In this section, we proceed to show how to estimate accurate correspondence labels for $\Tilde{\mathcal{D}}_n$ by using the anchor points $\Tilde{\mathcal{D}}_c$ collected by Eq.~\ref{eq:anchor_select}. The key idea of our method is based on a rational assumption that \textit{similar images should have similar textual descriptions and vice versa.} In other words, in an image-text shared feature space, if two images $I_1$ and $I_2$ are very close but their corresponding texts $T_1$ and $T_2$ are far away from each other, we can tell that at least one of these two data pairs (\textit{i.e.,} $(I_1,T_1)$ or $(I_2,T_2)$) is mislabeled. Furthermore, if we assume the $(I_1,T_1)$ is clean (well-matched), then the gap between $D(f(I_2),g(I_1))$ and $D(f(T_2),g(T_1))$ can reflect the degree of correlation between $I_2$ and $T_2$ to some extent, where $D(\cdot,\cdot)$ is a distance function in feature space and we will write $D(f(I),g(I))$ as $D(I,I)$ in the following. That is to say, if the distances between $(I_2,I_1)$ and $(T_2,T_1)$ are more \textit{consistent}, then $I_2$ and $T_2$ are more correlated.

Formally, for the $i$-th noisy data pair $(I_n^i,T_n^i)$ in $\Tilde{\mathcal{D}}_n$, we first search its closest image $I_c^\vartriangle$ in the collected anchor point set $\Tilde{\mathcal{D}}_c$, then we can compute its \textit{image2text similarity consistency} by comparing the image feature distance $D(I_n^i,I_c^\vartriangle)$ and their corresponding text feature distance $D(T_n^i,T_c^\vartriangle)$:
\begin{equation}
    \mathcal{C}_{i2t} = \frac{D(I_n^i,I_c^\vartriangle)}{D(T_n^i,T_c^\vartriangle)}, (I_n^i,T_n^i) \in \Tilde{\mathcal{D}}_n, (I_c^\vartriangle,T_c^\vartriangle) \in \Tilde{\mathcal{D}}_c
\end{equation}
Similarly, we can compute its \textit{text2image similarity consistency} by:
\begin{equation}
    \mathcal{C}_{t2i} = \frac{D(T_n^i,T_c^\lozenge)}{D(I_n^i,I_c^\lozenge)}, (I_n^i,T_n^i) \in \Tilde{\mathcal{D}}_n, (I_c^\lozenge,T_c^\lozenge) \in \Tilde{\mathcal{D}}_c
\end{equation}

where $T_c^\lozenge$ is the closest text feature to $T_n^i$ in the collected anchor point set $\Tilde{\mathcal{D}}_c$. The estimated soft correspondence label $y_i^*$ of $(I_n^i,T_n^i)$ is finally formulated by the bidirectional cross-modal similarity consistency:
\begin{equation}\label{eq:soft_label}
    y_i^* = (\frac{D(I_n^i,I_c^\vartriangle)}{D(T_n^i,T_c^\vartriangle)} + \frac{D(T_n^i,T_c^\lozenge)}{D(I_n^i,I_c^\lozenge)})/2
\end{equation}
Then the noisy data with the estimated soft correspondence labels $\Tilde{\mathcal{D}}_n = \{I_n,T_n, y_n = y^*\}$ and the clean data $\Tilde{\mathcal{D}}_c = \{I_c,T_c,y_c = 1 \}$ can be combined to train the matching model by minimizing the soft triplet loss in Eq.~\ref{eq:soft_loss}. The detailed training pipeline is illustrated in the Algorithm.~\ref{algo}. 


\begin{table*}[!ht]
	\setlength{\abovecaptionskip}{0.cm}
	\setlength{\belowcaptionskip}{-0.cm}
	\newcommand{\tabincell}[2]{\begin{tabular}{@{}#1@{}}#2\end{tabular}}
	\setlength{\belowdisplayskip}{1cm}
	\centering
	\caption { Image-Text Retrieval on Flickr30K and MS-COCO 1K.}
	\resizebox{\textwidth}{!}{
		\begin{tabular}{c|c|ccc|ccc|c|ccc|ccc|c}
		\toprule[1.5pt]
		&&\multicolumn{7}{c|}{Flickr30K}&\multicolumn{7}{c}{MS-COCO}\\
		&&\multicolumn{3}{c|}{Image$\longrightarrow$Text}&\multicolumn{3}{c|}{Text$\longrightarrow$Image}&&\multicolumn{3}{c|}{Image$\longrightarrow$Text}&\multicolumn{3}{c|}{Text$\longrightarrow$Image}&\\
		\hline
			Noise&Methods&R@1&R@5&R@10&R@1&R@5&R@10&Sum&R@1&R@5&R@10&R@1&R@5&R@10&Sum\\
			\midrule
			\multirow{9}{*}{20\%}&SCAN& 58.5     & 81.0   & 90.8 & 35.5 & 65.0   & 75.2 & 406.0   & 62.2 & 90.0   & 96.1 & 46.2 & 80.8 & 89.2 & 464.5\\
			~&VSRN& 33.4     & 59.5 & 71.3 & 25.0   & 47.6 & 58.6 & 295.4 & 61.8 & 87.3 & 92.9 & 50.0   & 80.3 & 88.3 & 460.6 \\
			~&IMRAM & 22.7     & 54.0   & 67.8 & 16.6 & 41.8 & 54.1 & 257.0   & 69.9 & 93.6 & 97.4 & 55.9 & 84.4 & 89.6 & 490.8 \\
			~&SAF & 62.8     & 88.7 & 93.9 & 49.7 & 73.6 & 78.0   & 446.7 & 71.5 & 94.0   & 97.5 & 57.8 & 86.4 & 91.9 & 499.1 \\
			~&SGR& 55.9     & 81.5 & 88.9 & 40.2 & 66.8 & 75.3 & 408.6 & 25.7 & 58.8 & 75.1 & 23.5 & 58.9 & 75.1 & 317.1 \\
			~&NCR & 73.5     & 93.2 & 96.6 & 56.9 & 82.4 & 88.5 & 491.1 & 76.6 & 95.6 & 98.2 & 60.8 & 88.8 & 95.0   & 515.0   \\
			~&DECL&77.5&93.8 &97.0&56.1&81.8&88.5&494.7&77.5&95.9&98.4&61.7&89.3&95.4&518.2\\
			~&\textbf{BiCro}&\textbf{78.3}&94.1&97.3&60.0&83.7&89.5&502.9&78.2&95.9&98.4&62.5&89.8&95.5&520.3\\
			~&\textbf{BiCro*}&78.1&\textbf{94.4}&\textbf{97.5}&\textbf{60.4}&\textbf{84.4}&\textbf{89.9}&\textbf{504.7}&\textbf{78.8}&\textbf{96.1}&\textbf{98.6}&\textbf{63.7}&\textbf{90.3}&\textbf{95.7}&\textbf{523.2}\\
			\midrule
			\multirow{9}{*}{40\%}&SCAN& 26.0       & 57.4 & 71.8 & 17.8 & 40.5 & 51.4 & 264.9 & 42.9 & 74.6 & 85.1 & 24.2 & 52.6 & 63.8 & 343.2 \\
			~&VSRN& 2.6      & 10.3 & 14.8 & 3.0    & 9.3  & 15.0   & 55.0    & 29.8 & 62.1 & 76.6 & 17.1 & 46.1 & 60.3 & 292.0   \\
			~&IMRAM& 5.3      & 25.4 & 37.6 & 5.0    & 13.5 & 19.6 & 106.4 & 51.8 & 82.4 & 90.9 & 38.4 & 70.3 & 78.9 & 412.7 \\
			~&SAF & 7.4      & 19.6 & 26.7 & 4.4  & 12.2 & 17.0   & 87.3  & 13.5 & 43.8 & 48.2 & 16.0   & 39.0   & 50.8 & 211.3 \\
			~&SGR& 4.1      & 16.6 & 24.1 & 4.1  & 13.2 & 19.7 & 81.8  & 1.3  & 3.7  & 6.3  & 0.5  & 2.5  & 4.1  & 18.4  \\
			~&NCR & 68.1     & 89.6 & 94.8 & 51.4 & 78.4 & 84.8 & 467.1 & 74.7 & 94.6 & 98.0   & 59.6 & 88.1 & 94.7 & 509.7 \\
			~&DECL&72.7&92.3&95.4&53.4&79.4&86.4&479.6&75.6&95.5&98.3&59.5&88.3&94.8&512.0\\
			~&\textbf{BiCro}&73.6&\textbf{93.0}&\textbf{96.4}&\textbf{56.0}&80.8&87.4&487.2&76.4&95.2&\textbf{98.6}&61.5&\textbf{89.4}&\textbf{95.5}&516.6\\
			~&\textbf{BiCro*}&\textbf{74.6}&92.7&96.2&55.5&\textbf{81.1}&\textbf{87.4}&\textbf{487.5}&\textbf{77.0}&\textbf{95.9}&98.3&\textbf{61.8}&89.2&94.9&\textbf{517.1}\\
			\midrule
			\multirow{9}{*}{60\%}&SCAN& 13.6     & 36.5 & 50.3 & 4.8  & 13.6 & 19.8 & 138.6 & 29.9 & 60.9 & 74.8 & 0.9  & 2.4  & 4.1  & 173.0   \\
			~&VSRN & 0.8      & 2.5  & 5.3  & 1.2  & 4.2  & 6.9  & 20.9  & 11.6 & 34.0   & 47.5 & 4.6  & 16.4 & 25.9 & 140.0   \\
			~&IMRAM & 1.5      & 8.9  & 17.4 & 1.9  & 5.0    & 7.8  & 42.5  & 18.2 & 51.6 & 68.0   & 17.9 & 43.6 & 54.6 & 253.9 \\
			~&SAF  & 0.1      & 1.5  & 2.8  & 0.4  & 1.2  & 2.3  & 8.3   & 0.1  & 0.5  & 0.7  & 0.8  & 3.5  & 6.3  & 11.9  \\
			~&SGR & 1.5      & 6.6  & 9.6  & 0.3  & 2.3  & 4.2  & 24.5  & 0.1  & 0.6  & 1.0    & 0.1  & 0.5  & 1.1  & 3.4   \\
			~&NCR  & 13.9     & 37.7 & 50.5 & 11.0   & 30.1 & 41.4 & 184.6 & 0.1  & 0.3  & 0.4  & 0.1  & 0.5  & 1.0    & 2.4   \\
			~&DECL&65.2&88.4&94.0&46.8&74.0&82.2&450.6&73.0&94.2&\textbf{97.9}&57.0&86.6&93.8&502.5\\
			~&\textbf{BiCro}&\textbf{68.3}&90.4&93.8&\textbf{51.9}&76.9&84.4&465.7&\textbf{73.9}&\textbf{94.7}&97.7&\textbf{58.7}&87.0&93.8&\textbf{505.8}\\
			~&\textbf{BiCro*}&67.6&\textbf{90.8}&\textbf{94.4}&51.2&\textbf{77.6}&\textbf{84.7}&\textbf{466.3}&\textbf{73.9}&94.4&97.8&58.3&\textbf{87.2}&\textbf{93.9}&505.5\\
			
			
			\bottomrule[1.5pt]
	\end{tabular}}
	\label{table:flicker}
\vspace{-1mm}
\end{table*}

\section{Experiment}
In this section, we evaluate the effectiveness of our proposed method \textit{BiCro} on three image-text matching datasets, \textit{i.e.,} Flickr30K~\cite{young2014image}, MS-COCO~\cite{mscoco}, and Conceptual Captions~\cite{sharma2018conceptual}. The Flickr30K~\cite{young2014image} and MS-COCO~\cite{mscoco} and two well-annotated datasets, we randomly corrupt their image-text data pairs for a specific percentage (\textit{i.e.,} noise ratio) to simulate the noisy correspondence issue. The Conceptual Captions~\cite{sharma2018conceptual} is with real noisy correspondence from the wild.

\subsection{Datasets and Evaluation Metrics}
The following three widely-used image-text matching datasets are used to evaluate our method and baselines:

\noindent \textbf{Flickr30K~\cite{young2014image}} contains 31,000 images collected from the Flickr website and each image is associated with five captions. We use 1,000 images for model validation, 1,000 images for model testing, and 29,000 for model training.

\noindent \textbf{MS-COCO~\cite{mscoco}} has 123,287 images with 5 captions each. Among them, 5,000 images are used for modal validation, 5,000 images are used for model testing, and 113,287 images are used for model training.

\noindent \textbf{Conceptual Captions~\cite{sharma2018conceptual}} is a large-scale image-text dataset with real-world noisy correspondence problem. It contains 3.3M images and each image is associated with one caption. All the data pairs in the Conceptual Captions dataset are automatically harvested from the Internet, therefore about 3\%$\sim$20\% image-text pairs in the dataset are mismatched or weakly-matched~\cite{sharma2018conceptual}. Following NCR~\cite{NCR}, we use a subset of the Conceptual Captions dataset, \textit{i.e.}, CC152K, in our experiments. 150,000 images in CC152K are used for model training, 1,000 images are used for model validation, and 1,000 images are used for model testing.

\subsection{Implementation Details}
As a general framework, \textit{BiCro} can be applied to many existing cross-modal matching methods. Same as NCR~\cite{NCR} and DECL~\cite{QinP00022}, we implement \textit{BiCro} based on SGRAF~\cite{SGRAF}, whose performance is the state-of-the-art in image-text matching.
We follow the same training settings (\textit{e.g.,} optimizer, network architecture, all the hyperparameters, etc.) as previous works~\cite{NCR,QinP00022} to make a fair comparison. Please refer to NCR~\cite{NCR} for more training 
details.

We first warmup the matching models $A$ and $B$ for 10 epochs to make them achieve an initial convergence. To reduce the effect of noisy data pairs, we select a small portion of data with small loss value in each batch to warmup the models based on a predefined warmup selection ratio $\alpha$. The choice of $\alpha$ will be discussed in Section.~\ref{sec:ablation}.
At the training stage, the total number of iterations is 40 epochs, among them, the first 20 epochs are trained with clean samples screened by BMM, and the next 20 epochs were trained on all samples (with our estimated soft labels). At each training epoch, we select 10\% data pairs with a higher probability of being clean in Eq.~\ref{eq:anchor_select} as the selected anchor points. The remained data are regarded as noisy ones. At the inference stage, we average the similarities predicted by networks $A$ and $B$ for the retrieval evaluation. 
We propose two strategies to handle the noisy data pairs, (1) feed all of the data pairs with their estimated soft labels to the matching model (denoted as BiCro in Table.~\ref{table:flicker} and Table.~\ref{table:cc152k}), (2) treat those data pairs with estimated soft labels under a threshold (mismatch threshold $\theta$) as \textit{mismatched data}, and set their correspondence labels as zero (denoted as BiCro* in Table.~\ref{table:flicker} and Table.~\ref{table:cc152k}). The effect of the mismatch threshold $\theta$ will be discussed in Section.~\ref{sec:ablation}.

\subsection{Comparison with State-of-the-Arts}
To demonstrate the effectiveness of the proposed \textit{BiCro}, we evaluate the \textit{BiCro} against several baseline methods, including general methods (SCAN~\cite{SCAN}, VSRN~\cite{VSRN}, IMRAM~\cite{imram}, SGRAF, SGR and SAF~\cite{SGRAF}) and robust learning method against noisy correspondence ( NCR~\cite{NCR} and DECL~\cite{QinP00022} ). 


\subsubsection{Results on Simulated Noise}
Table.~\ref{table:flicker} reports the experimental results on the 1K test images of Flickr30K dataset and over 5 folds of 1K test images of MS-COCO dataset. From the experimental results, we can find that the proposed \textit{BiCro} performs significantly better than the state-of-the-art methods. In comparison to the best baseline DECL~\cite{QinP00022}, BiCro improves the sum score for retrieving by 8.2\%,7.6\%,15.1\%,2.1\%,4.6\% and 3.3\% under different noise rates, respectively. In addition, BiCro* further improves the overall performance by filtering out the mismatched pairs. The results at a noise rate of 60\% show a strong robustness of our \textit{BiCro} against the noisy correspondence with high noise rates. 
The failure of NCR to deal with high noise rates may be that it cannot divide the noisy and clean pairs well relying on GMM \cite{NCR}, while our method solves this drawback by using BMM.

\begin{table}[t!]
	\setlength{\abovecaptionskip}{0.cm}
	\setlength{\belowcaptionskip}{0.cm}
	\makeatletter\def\@captype{table}
	\caption{Image-Text Retrieval on CC152K.}
	\resizebox{\linewidth}{!}{
		\begin{tabular}{c|ccc|ccc|c}
			\toprule[1.5pt]
			& \multicolumn{3}{c|}{Image$\longrightarrow$Text}&\multicolumn{3}{c|}{Text$\longrightarrow$Image}&\\
			\hline
			Methods&R@1&R@5&R@10&R@1&R@5&R@10&Sum\\
			\hline
			SCAN&30.5&55.3&65.3&26.9&53.0&64.7&295.7\\
			VSRN&32.6&61.3&70.5&32.5&59.4&70.4&326.7\\
			IMRAM&33.1&57.6&68.1&29.0&56.8&67.4&312.0\\
			SAF&31.7&59.3&68.2&31.9&59.0&67.9&318.0\\
			SGR&11.3&29.7&39.6&13.1&30.1&41.6&165.4\\
			NCR&39.5&64.5&73.5&40.3&64.6&73.2&355.6\\
			DECL&39.0&66.1&75.5&40.7&66.3&76.7&364.3\\
			\textbf{BiCro}&40.7&\textbf{67.3}&\textbf{76.7}&39.7&\textbf{67.6}&\textbf{76.9}&368.9\\
			\textbf{BiCro*}&\textbf{40.8}&67.2&76.1&\textbf{42.1}&\textbf{67.6}&76.4&\textbf{370.2}\\

			\bottomrule[1.5pt]
		\end{tabular}
	}
	\label{table:cc152k}
 \vspace{-3mm}

\end{table}
\begin{table}[t!]
\centering
	\setlength{\abovecaptionskip}{0.cm}
	\setlength{\belowcaptionskip}{0.cm}
	\makeatletter\def\@captype{table}
	\caption{Ablation studies on Flickr30K with 40\% noise.}
	\resizebox{\linewidth}{!}{
		\begin{tabular}{cccc|ccc|ccc}
			\toprule[1.5pt]
			\multicolumn{4}{c|}{Methods}& \multicolumn{3}{c|}{Image$\longrightarrow$Text}&\multicolumn{3}{c}{Text$\longrightarrow$Image}\\
			\hline
			\thead{Co- \\ teaching}&\thead{Soft \\ label}&Bmm&Warmup&R@1&R@5&R@10&R@1&R@5&R@10\\
			\bottomrule[0.5pt]
			\checkmark&\checkmark&\checkmark&\checkmark&\textbf{74.6}&\textbf{92.7}&\textbf{96.2}&\textbf{55.5}&\textbf{81.1}&\textbf{87.4}\\
			&\checkmark&\checkmark&\checkmark&69.9&91.9&95.7&52.0&79.0&85.5\\
			\checkmark&&\checkmark&\checkmark&72.0&91.0&95.0&53.3&78.8&86.1\\
			\checkmark&\checkmark&&\checkmark&72.0&92.3&95.5&55.2&79.8&86.6\\
			\checkmark&\checkmark&	\checkmark&&54.3&81.9&88.1&40.9&65.8&73.6\\
			
			\bottomrule[1.5pt]
		\end{tabular}
	}
	\vspace{-4mm}
	
	\label{table:ablation}
\end{table}
\subsubsection{Results on Real-world Noise}
We evaluate the proposed method under the real noisy correspondence of CC152K. The experimental results are reported in Table.~\ref{table:cc152k}. From the results, one could observe that our \textit{BiCro} could achieve competitive performance under real noise. Specifically, \textit{BiCro} is 4.6\% higher than the best baseline in terms of sum in retrieval, respectively. Moreover, the performance gap between \textit{BiCro} and \textit{BiCro*} shows that the filtering of data pairs according to soft correspondence labels can further reduce the impact of data mismatch issue on performance.

\subsection{Hyperparameter Analysis}

\begin{figure}[t!]
\centering
  \resizebox{0.9\linewidth}{!} {
    \includegraphics{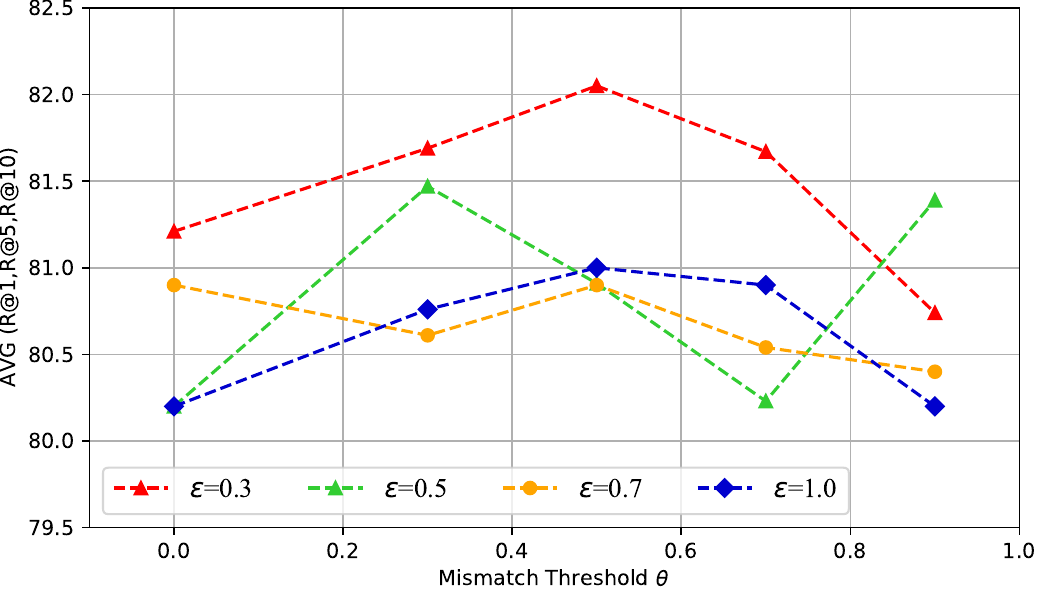}
  }
  \caption{Variation of retrieval performance with different selection ratio $\epsilon$ and mismatch threshold $\theta$. Note that $\theta$ = 0 denotes using the original generated soft labels for all data pairs.}
  \vspace{-5mm}
    \label{fig:hyperparameter}
\end{figure}

We now analyze the effect of the warmup selection ratio $\epsilon$ and mismatch threshold $\theta$, which denote the ratio of data that participated in warmup and the threshold that sets the soft label to 0. We report the average performance of retrieving image and text with different $\epsilon$ and $\theta$ on Flickr30K with the noise ratio of 40\%. As shown in Fig.~\ref{fig:hyperparameter}, setting $\epsilon=0.3$ achieves the best overall accuracy for retrieving. Using all data ($\epsilon=1$) in warmup gives limited performance, which indicates it is necessary to select low loss data in warmup. On the other hand, with the increases of $\theta$, our \textit{BiCro*} improves the accuracy compared with using the original generated soft labels ($\theta=0$). Moreover, further increasing the $\theta$ leads to the performance drop since a too-large mismatch threshold $\theta$ treats weakly labeled samples as mislabeled samples, resulting in the decline of generalization ability.

\subsection{Ablation Study}\label{sec:ablation}
In this section, we carry out the ablation study on the Flickr30K with a noise ratio of 40\%. We show the effect of each component in Table.~\ref{table:ablation}. In order to prob the effectiveness of soft label estimation in \textit{BiCro}, the labels of clean and noisy samples divided by co-teaching are set to 1 and 0 respectively as a comparison experiment to \textit{BiCro} and the experimental results are shown in the third row. The fourth row shows the experiments of our method with GMM and BMM. From the results, we draw the following conclusions: 1) Since the quality of noise-robust cross-modal learning depends on the soft correspondence labels, the performance of adopting the soft labels estimated by \textit{BiCro} surpasses the performance of using the original binary labels. 2) Although leveraging Gaussian-Mixture-Model achieves decent results, the performance can be further improved by Beta-Mixture-Model. 3) Co-teaching and Warmup, as important modules of the base network, are also crucial in our framework BiCro. 4) The model achieves the best test accuracy by utilizing all the components, which shows the complementarity of our proposed \textit{BiCro}.

\section{Conclusion}
This paper focuses on the challenge of robust cross-modal matching on noisy data. To address this problem, We propose a general framework called \textit{bidirectional cross-modal similarity consistency (BiCro)} for soft correspondence label estimation given only noisily-collected data pairs. The effectiveness of the proposed framework was verified on both synthetic noisy datasets and real noisy dataset. The visualization results also demonstrate that our estimated soft labels are the accurate estimation of the true correlation degree between different modality data.

\section{Acknowledgements}
This work was partially supported by the National Key R\&D Program of China (No.2021ZD0110901).

{\small
\bibliographystyle{ieee_fullname}
\bibliography{egbib}
}

\end{document}